\documentclass{article}

\PassOptionsToPackage{numbers, compress}{natbib}

\usepackage[preprint]{neurips_2022}

\usepackage[resetlabels]{multibib}
\newcites{App}{Appendix References}
\usepackage{minitoc}

\usepackage{algorithm}%
\usepackage{algorithmic}

\usepackage[utf8]{inputenc} %
\usepackage[T1]{fontenc}    %

\usepackage{url}            %
\usepackage{booktabs}       %
\usepackage{amsfonts}       %
\usepackage{nicefrac}       %
\usepackage{microtype}      %
\usepackage{comment}
\usepackage{xcolor}         %
\usepackage{physics}
\usepackage{bbm}
\usepackage{caption}
\usepackage[export]{adjustbox}
\usepackage[position=bottom]{subfig}
\usepackage{graphicx}
\newcommand{\bx}{\mathbf{x}}
\newcommand{\bz}{\mathbf{z}}
\usepackage{setspace}
\usepackage{mathtools}
\usepackage{array, makecell, multirow}
\usepackage{bm}
\DeclarePairedDelimiter{\ceil}{\lceil}{\rceil}
\everypar{\looseness=-1}
\newcolumntype{M}[1]{>{\centering\arraybackslash}m{#1}}

\usepackage{tikz}
\usetikzlibrary{shapes}
\usetikzlibrary{positioning}
\usetikzlibrary{plotmarks}
\usetikzlibrary{patterns}
\usetikzlibrary{intersections,shapes.arrows}
\usepackage{pgfplots}
\usetikzlibrary{pgfplots.fillbetween}
\definecolor{royalazure}{rgb}{0.0, 0.22, 0.66}
\usepackage[colorlinks = true,
            linkcolor = black,
            urlcolor  = black,
            citecolor = black,
            anchorcolor = black]{hyperref}

\title{Triangular Contrastive Learning on Molecular Graphs}

\author{%
  MinGyu Choi$^{1, 2}$ \\
  \And
  Wonseok Shin$^{1, 2}$ \\
  \And
  Yijingxiu Lu$^{1}$
  \And
  Sun Kim$^{1, 3}$\\ 
  \AND\vspace{-0.7cm}\\
  \texttt{\{chemgyu98, gratus907, solanoon, sunkim.bioinfo\}}\text{@snu.ac.kr}  \vspace{0.1cm}\\
  $^1$Seoul National University \\
  $^2$AIGENDRUG Co. Ltd \\
  $^3$MOGAM Institute for Biomedical Research 
}

\begin{document}

\doparttoc %
\faketableofcontents %

\maketitle
\begin{abstract}
Recent contrastive learning methods have shown to be effective in various tasks, learning generalizable representations invariant to data augmentation thereby leading to state of the art performances. Regarding the multifaceted nature of large unlabeled data used in self-supervised learning while majority of real-word downstream tasks use single format of data, a multimodal framework that can train single modality to learn diverse perspectives from other modalities is an important challenge.
In this paper, we propose TriCL (Triangular Contrastive Learning), a universal framework for trimodal contrastive learning.
TriCL takes advantage of \textit{Triangular Area Loss}, a novel intermodal contrastive loss that learns the angular geometry of the embedding space through simultaneously contrasting the area of positive and negative triplets.
Systematic observation on embedding space in terms of alignment and uniformity showed that Triangular Area Loss can address the \textit{line-collapsing problem} by discriminating modalities by angle.
Our experimental results also demonstrate the outperformance of TriCL on downstream task of molecular property prediction which implies that the advantages of the embedding space indeed benefits the performance on downstream tasks.
\end{abstract}

\section{Introduction}
\label{introduction}
Data scarcity has been a severe problem in representation learning, due to the time-consuming and high-cost nature of annotating large-scale data \citep{TLSurvey}. In the field of self-supervised learning (SSL), contrastive learning (CL) that learns the general landscape of an embedding space from unlabeled data by pulling similar (\textbf{positive}) pairs and pushing dissimilar (\textbf{negative}) pairs \citep{jaiswal2020survey, jing2020self}, have shown promising strength in learning diverse characteristics from multiple viewpoints without labels when compared with traditional supervised learning methods. Multimodal CL is especially powerful for the data whose characteristics are naturally hard to be expressed comprehensively with individual representation \citep{clip, yuan2021multimodal, zolfaghari2021crossclr}. 

As most of the real-world downstream tasks use data from a single modality in fine-tuning, generating informative embedding space that can be fully utilized by a single encoder (referred to as \textbf{main encoder}) of the multimodal CL framework becomes extremely important. The special characteristics of multimodal CL require views from other modalities (referred to as \textbf{auxiliary modalities}) to be distilled and mapped into the embedding space. Meanwhile, many works recently started introducing more modalities into CL \citep{hyconTri} while existing objectives were mainly proposed for unimodal or bimodal networks \citep{InfoNCE, NTXent}. Thus, design of scalable framework and appropriate contrastive objective for higher-modality is urgently needed.

To address the two challenges described above, we introduce TriCL, a trimodal contrastive learning framework with novel Triangular Area Loss. TriCL focuses on fully utilizing trimodal information to build an effective embedding space for downstream tasks. Our contributions can be summarized as:

\begin{enumerate}
    \item \textbf{Observation on the geometry of trimodal embedding space.} Expanding the analysis by \citet{MCLspace}, we characterize the embedding space produced by multimodal CL in terms of intermodal alignment and uniformity. We demonstrate that intramodal contrastive loss distributes the embedding space while the intermodal loss compresses.
    \item \textbf{Proposal of Triangular Area Loss, geometry-aware contrastive loss for trimodal CL.} We analyze the reasons behind space collapse in intermodal CL by readdressing the importance of intermodal uniformity. Triangular Area Loss is proposed and formulated under TriCL, a universal framework for trimodal CL. Triangular Area Loss takes a glimpse of geometry through minimizing and maximizing areas of triangles, instead of their pairwise distances. We also demonstrate that the embedding space produced by TriCL with our loss displays more useful properties than the space optimized with the pairwise loss objective.
    \item \textbf{State-of-the-Art performance on molecular property prediction tasks.} We proved the utility of TriCL by achieving the best AUC-ROC on most molecular property prediction tasks over the latest methods. Ablations on objective functions prove that geometric advantages in embedding space result in improvement of downstream task performance.
\end{enumerate}

\section{Related Works}
\label{relatedworks}

Contrastive learning (CL) has recently shown the competitive power of learning transferable knowledge from large-scale unlabeled data for downstream tasks \citep{Islam2021Transferability}. Building effective representations that are invariant from different views of data is one of the most important missions in CL that demands the minimization of irrelevant nuisances during the pretraining process \citep{GoodView}. \citet{PropCL} emphasized clever choice of the contrastive loss function and augmentation strategy.

\citet{TripletLoss} designs the \textbf{Triplet Margin Loss} by setting an anchor as the criterion for pulling positive samples and pushing negative ones. Based on information theoretic arguments, \citet{InfoNCE} develops a loss function named \textbf{InfoNCE} to optimize the lower bound of mutual information between the encoded representations. \textbf{NT-Xent (Normalized Temperature-scaled Cross Entropy)} suggested by \citet{NTXent} compares multiple negatives effectively to identify positives. Later, \citet{SimCLR} leverages the normalization and temperature of NT-Xent loss which performs best on their visual representation learning framework. 

To measure the strength of contrastive objective in downstream tasks, \citet{MCLspace} characterized two desirable properties in the following: 
\begin{itemize}
\setlength{\itemindent}{-2em}
  \item \textbf{Alignment}: Positive pairs are mapped closely in the embedding space.
  \item \textbf{Uniformity}: Embeddings are uniformly distributed, preserving as much information as possible.
\end{itemize}
\citet{PropCL} gives the theoretical argument of the significance of alignment and uniformity similarly.

The multimodal characteristics of data have inspired researches on multimodal CL in computer vision domain. To jointly train the image encoder and the text encoder, \citet{clip} calculates the cosine similarity between the embeddings of image and text for all pairs across a batch. \citet{yuan2021multimodal} takes both intra and intermodal similarities into account, enforcing consistency within modality and introducing influential samples from another modality simultaneously to preserve the semantic similarity. 
Focused on learning a cross-modal embedding, \citet{zolfaghari2021crossclr} introduces CrossCLR loss to further ensure the intramodal proximity for improving cross-modal retrieval performance. 

When it comes to graph contrastive learning, \citet{GraphCL} proposes several graph augmentations including node dropping and masking, and notes the importance of augmentation selection for different tasks. \citet{ADGCL} demonstrates the importance of avoiding capturing redundant information to identify graphs in contrastive learning. Meanwhile, molecules as commonly used graph benchmark data have also become the research highlights of graph representation learning. To thoroughly represent the molecular information, \citet{MolCLR} builds a framework for learning molecular graph representations by contrasting positives against negatives. \citet{graphmvp} includes a 3D representation of molecules as an additional modality to consider stochasticity for capturing the conformer distribution of a 2D graph.

\section{Observations and Explanations on Trimodal Embedding Space.}
\label{observation}
\begin{figure}
  \centering
    \includegraphics[width=\linewidth]{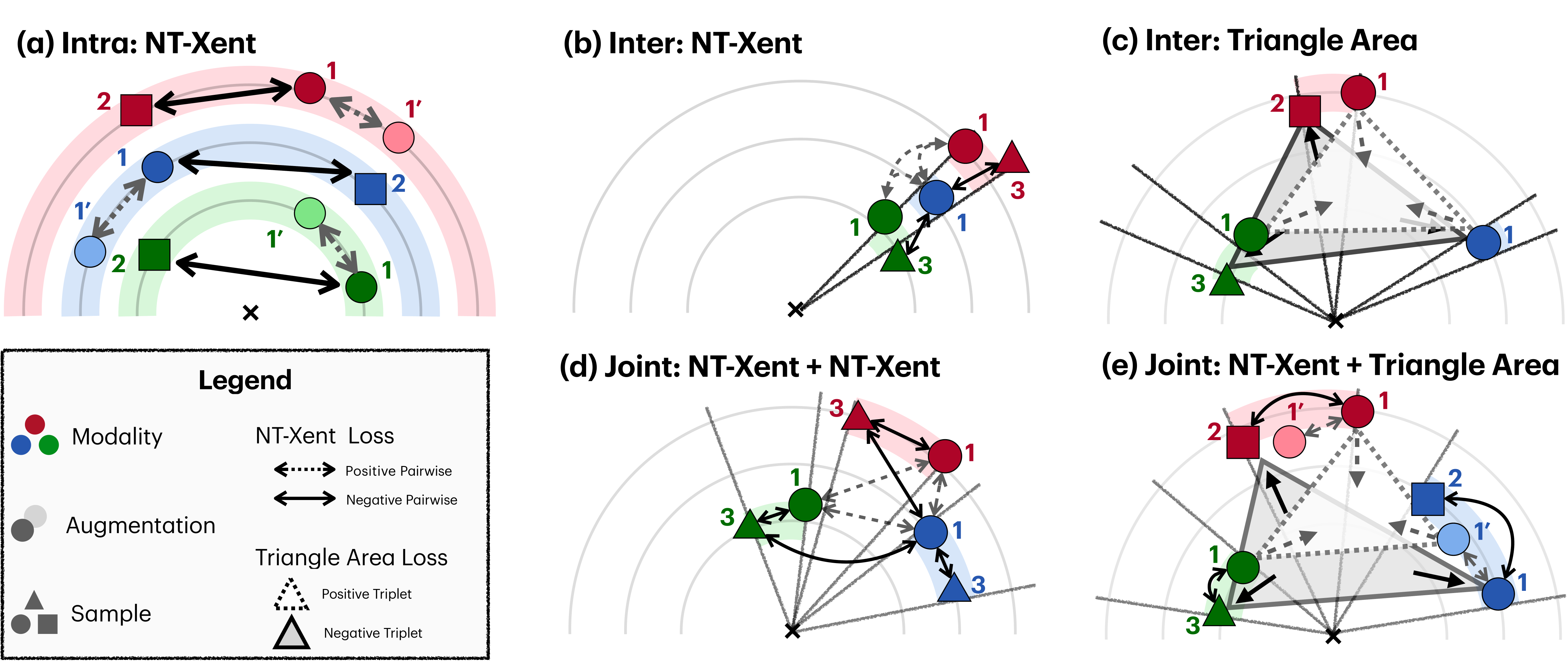}
  \caption{Illustration of embedding space after trimodal contrastive learning. Specific loss function and geometry of each space: (a) NT-Xent as intramodal loss: `hypersphere' (b) NT-Xent as intermodal loss: `line' (c) Triangle Area Loss as intermodal loss: `line' (d) NT-Xent as intra- and intermodal loss: `cone' (e) Triangle Area Loss as intermodal loss, NT-Xent as intramodal loss: `cone'. Angles within the space and angles between them are not to scale. Refer to the Table \ref{alignment_metric} for quantified metrics.}
  \label{fig:fig1}
\end{figure}

Inspired by the bipartite components of NT-Xent loss which quantitatively pulls positive pairs and pushes negative pairs, we start by inspecting how alignment and uniformity as contrastive loss optimize the joint embedding space. We first implement a simple trimodal framework comprising transformer, GNN, and 3D CNN to encode text, graph, and structure of molecules respectively. 

\subsection{Alignment and Uniformity in Multimodal Contrastive Learning}
\label{3.1}
Analogous to the alignment and uniformity in unimodal NT-Xent loss, we expand the scope of discussion into the multimodal CL by introducing the concept of intermodal alignment and intermodal uniformity (\autoref{eq:Lmulti2}). An intermodal alignment regulates to what extent an encoder learns sample diverse perspectives from other encoders and generates representations in regard to multiple views. Conversely, an intermodal uniformity enhances encoder discriminability in capturing distinct features that are unobservable in one modality by contrasting them with negatives from other modalities.
\begin{equation}
\label{eq:Lmulti2}
\mathcal{L} = ( \mathcal{L}_{\mathrm{intra}}^{\mathrm{align}} + \mathcal{L}_{\mathrm{intra}}^{\mathrm{uniform}}) + \lambda_{\mathrm{inter}}( \mathcal{L}_{\mathrm{inter}}^{\mathrm{align}} + \mathcal{L}_{\mathrm{inter}}^{\mathrm{uniform}})
\end{equation}

In fact, intramodal uniformity and intermodal alignment are not independent. When samples result in distinct representations on the main encoder but similar on the auxiliary modalities, intermodal alignment would pull these representations closer with the expense of intramodal uniformity. Regarding that multimodal CL aims to train the main encoder to reflect similarities from auxiliary representations, the balance between intramodal uniformity and intermodal alignment would be the key objective for successful multimodal CL.

\begin{table}
  \caption{Metrics regarding the embedding space after trimodal contrastive learning. Alignment metric is the average cosine similarity between all positive pairs (higher is better). Uniformity metric is the average cosine similarity between randomly chosen pairs (close to 0 is better). Combined metric refers to $\mathrm{(Align - \abs{Uniform})}$ (higher is better). For triplets, all metrics are computed as the average pairwise metric. NT-Xent loss uses temperature $\tau$ = 0.1. For implementation details, see Appendix D.}
  \label{alignment_metric}
  \centering
  \small  
  \begin{tabular}{cc|rrr|rrr}
    
    \toprule
    \multicolumn{2}{c|}{\multirow{2}{*}{Loss}} & \multicolumn{3}{c|}{Intramodal (Main encoder)} & \multicolumn{3}{c}{Intermodal}
    \\\cmidrule(r){3-8}
    & & Align & Uniform & Combined & Align & Uniform & Combined \\ 
    \midrule
    Intra       & NT-Xent   & 0.663 & 0.001 & 0.662 & -0.003 & 0.000 &-0.003 \\ 
    \midrule
    Inter       & NT-Xent   & 0.996 & 0.998 & -0.002 &  1.000 & 0.999  &0.001\\ 
                & Ours      & 0.997 & 1.000 & -0.003 & 0.002 & 0.002 & 0.000 \\ 
    \midrule
    Joint       & NT-Xent   & 0.660 & 0.036 & 0.624 &0.101 & 0.091  & 0.010\\ 
                & Ours      & 0.694 & 0.004 & \textbf{0.690} &0.138 & 0.079 & \textbf{0.049}\\
    \bottomrule
  \end{tabular}
\end{table}

\subsection{Transformation of an embedding space affected by Intra and Intermodal NT-Xent loss}
\label{3.2}

We design two experiments to explore how the embedding space transforms when optimized with contrastive loss. Specifically, encoders are pre-trained twice under different combinations of two losses: (1) Intramodal loss between augmented data within modality. (2) Intermodal loss between pairs of three modalities. At this time, only NT-Xent loss is applied to both experiments.

We observed the transformations of embedding spaces and measured the contributions of each loss component through two cosine similarity metrics (\autoref{alignment_metric}):
Intramodal cosine similarities between positive pairs and negative pairs are calculated respectively to assess the effects of intramodal alignment and intramodal uniformity.
Similarly, the sum of intermodal pairwise cosine similarities of positive triplets and negative triplets are calculated respectively to reflect contributions of intermodal alignment and intermodal uniformity. 
Upon the results, we draw the conceptual prospect of joint embedding spaces under each experiment setting in \autoref{fig:fig1}(a, b). 

\paragraph{Intramodal contrastive loss distributes the embedding space to a hypersphere.} 

Applying intramodal NT-Xent loss as an exclusive loss within modality is identical to the CL of three independent modalities.
We observe that under this setting, the intramodal alignment metric keeps increasing during the training process until it converges to 0.663.
Intramodal uniformity metric close to zero indicates that the encoder could distinguish individual samples, within the scope of the specific encoder itself.
Low intermodal alignment metric implies that positive representations are randomly spread over the space, which is straightforward as no information is exchanged over different modalities.

\paragraph{Intermodal contrastive loss compresses the embedding space into a `line'.}

To assure that one encoder can borrow diverse viewpoints from extra modalities, we apply NT-Xent losses on inter-modal pairs over three encoders. 
The intramodal alignment metric of three encoders rapidly approaches 0.996, indicating the collapse of individual embedding spaces.
Interestingly, according to the 1.000 intermodal alignment, embedding spaces of three modalities mostly overlap each other resulting in a single line.

About zero intramodal combined metric also implies that the encoder is nearly unable to distinguish different representations, leading to the ineffectiveness of the joint embedding space.
This result is counter-intuitive, as the intermodal uniformity loss would separate individual embedding clusters apart.
To the best of our knowledge, we guess this line-shaped joint space is a local minimum easy to fall while utilizing NT-Xent loss for intermodal contrast.

\section{TriCL : Triangular Contrastive Learning}
\label{tricl}

In this section, we reconsider the intermodal uniformity as a regularization objective.
Based on this intuition, we would introduce Triangular Area Loss, which aims to learn the geometry of the embedding space thereby preventing the collapse we observed above.
Triangular Area Loss would be formulized under TriCL (Triangular Contrastive Learning), a universal framework for trimodal CL applying our objective.
Advantages of Trianglular Area Loss and TriCL would also be discussed in Section \ref{3.2} inheriting the same views of embedding space.

\subsection{Readdressing intermodal uniformity: Geometry-aware contrastive loss}
\label{4.1}
As discussed in Section \ref{3.1} and \citep{MCLspace, PropCL}, minimizing NT-Xent loss could be interpreted as optimizing embedding space that satisfies two desirable properties: alignment and uniformity.
In multimodal contrastive learning, a network should additionally consider intermodal alignment and uniformity.

Minimizing intermodal alignment to pull inter-modal positive pair closer is a straightforward way to guide main encoder for reflecting perspectives from auxiliary modalities.
However, diminishing intermodal uniformity seems counter-intuitive, as harmonized embedding space is commonly considered an ideal space whose embedding vectors from the same samples converge into a single point.

At this time, we would emphasize the role of intermodal uniformity as a regularization factor, which prevents encoders from falling into the local minima.
As we observed in Section \ref{3.2}, a premature utilization of NT-Xent loss across modalities results in a "line" space.
Although this shape of embedding space looks harmonized and aligned, it is clearly undesirable because this embedding cannot distinguish samples yet injudiciously collapse into a single "line".

We found the reason for collapse through imprudently applying intermodal NT-Xent into trimodal CL, which pushes and pulls representations without considering the geometry of the embedding space. 

Triangle Area Loss we devised explicitly considers the geometry among embedding vectors by calculating the area of a triangle with three representations.
Specifically, as an area of a triangle is calculated with two sides and the angle between, \textbf{the encoder becomes aware of the angle between two sides that reflects geometry}, which was unavailable when applying intermodal NT-Xent loss. This discourages collapse of the embedding space to a single `line' by explicitly enlarging angles between negative triplets, which is equivalent to optimization of intermodal uniformity.
\begin{figure}
  \centering
    \includegraphics[width=\linewidth]{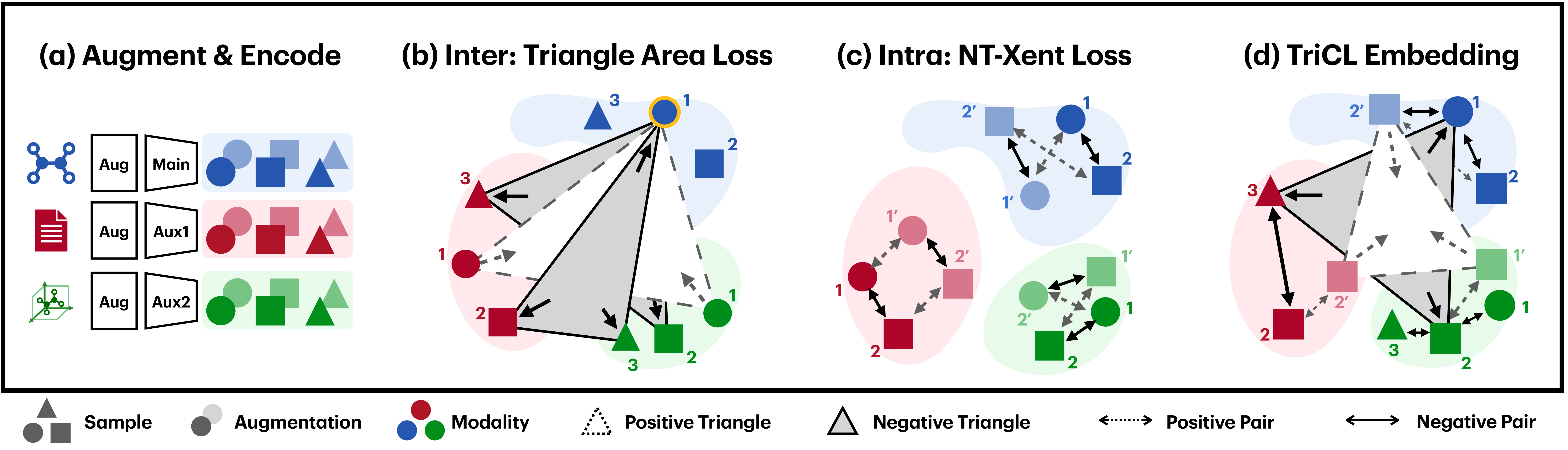}
  \caption{The TriCL framework. (a) Each sample is represented as three distinct format; after augmented twice then encoded generating six reprsentations per sample. (b) Representations in different modalities are contrasted using Triangle Area Loss. (c) Representations in the same modality are contrasted using pairwise NT-Xent loss. (d) TriCL build the embedding space by carefully balancing intramodal and intermodal contrastive loss.}
\end{figure}

\subsection{Architecture and Learning Objectives}
\label{4.2}
We would continue our discussions on Triangular Area Loss over TriCL, a universal trimodal contrastive learning framework appropriate for any types of pre-training using any network architectures and any data formats which can be decomposed to three modalities.

TriCL is designed to pre-train a \texttt{main} encoder with two other auxiliary modalities, \texttt{aux1} and \texttt{aux2}. 
By contrasting positive triplets and negative triplets at once, TriCL aims to train the \texttt{main} encoder to learn similarities and differences between inputs that are only recognizable for auxiliary modalities so that the \texttt{main} encoder can better capture general properties apt to downstream tasks.

\paragraph{Architecture}
Three encoder networks, \texttt{main, aux1, aux2} are given their respective set of inputs $\bm m^{\mathrm{main}}_j, \bm m^{\mathrm{aux1}}_j, \bm m^{\mathrm{aux2}}_j$ which are multiple views of the same sample $\bm m_j$. Each sample $\bm m^{\mathrm{enc}}_j$ is then augmented twice as $\bx^{\mathrm{enc}}_{2j-1}, \bx^{\mathrm{enc}}_{2j}$, following probabilistic augmentation strategies. The output of each encoder is a vector of equal length, defined as $\mathrm{enc}(\bx^{\mathrm{enc}}_{2j}) = \bz^{\mathrm{enc}}_{2j}$ where $\mathrm{enc}$ is one of \texttt{main, aux1, aux2}. Implementation details of augmentation is described in Appendix D.

\paragraph{Learning Objective}
As discussed in Section \ref{3.1}, a learning objective for multimodal contrastive learning should be carefully designed to maintain a balance between alignment and uniformity of intra and inter modalities. For this, the learning objectve of TriCL is formulized as a weighted sum of the intramodal contrastive loss with the intermodal contrastive loss.
\begin{equation*}
\mathcal{L} = ( \mathcal{L}_{\mathrm{intra}}^{\mathrm{align}} + \mathcal{L}_{\mathrm{intra}}^{\mathrm{uniform}}) + \lambda_{\mathrm{inter}}( \mathcal{L}_{\mathrm{inter}}^{\mathrm{align}} + \mathcal{L}_{\mathrm{inter}}^{\mathrm{uniform}})
\end{equation*}

For intermodal loss, TriCL adopts Triangular Area Loss as an objective. To formulize Triangle Area Loss, we first define \textbf{positive triplet} $\mathbf{P}$ as a triplet whose inputs are trimodal augmentations from the same sample, and the rest as \textbf{negative triplet} $\mathbf{N}$. 
Note that in a sample batch of size $B$, every sample is augmented twice, and thus there are $8B$ positive triplets and $8(B^3 - B)$ negative triplets. 

We then define the triangular contrastive metric as: 
\begin{equation} \label{triangularContMet}
    \mathcal{L}_{\mathrm{inter}} = \underbrace{\mathbb{E}\left[\mathrm{Area}(\bz^{\mathrm{main}}_i, \bz^{\mathrm{aux1}}_j, \bz^{\mathrm{aux2}}_k)^2 \mid  \mathbf{P}\right]}_{\text{\normalsize intermodal alignment}} - 
    \underbrace{\mathbb{E}\left[\mathrm{Area}(\bz^{\mathrm{main}}_i, \bz^{\mathrm{aux1}}_j, \bz^{\mathrm{aux2}}_k)^2 \mid \mathbf{N}\right]}_{\text{\normalsize intermodal uniformity}}
\end{equation}
In equation \ref{triangularContMet}, expectation is taken over all triplets of augmented data, and $\mathrm{Area}(\mathbf{a}, \mathbf{b}, \mathbf{c})$ refers to the area of triangle whose vertices are defined by $\mathbf{a}, \mathbf{b}$, and $\mathbf{c}$. Square is taken to reduce numerical instability in computation of triangular area. 

To compute on batch with size of $B$, triangular contrastive metric can be implemented as:
\begin{equation}
    \mathcal{L}_{\mathrm{inter}} = \sum_{(i, j, k) \in \left\{1,2,\dots 2B\right\}^3} \alpha\big(\ceil{i/2}, \ceil{j/2}, \ceil{k/2}\big) \cdot \mathrm{Area}(\bz^1_{i}, \bz^2_{j}, \bz^3_{k})^2
\end{equation}
Where $\alpha$ is a normalization factor accounting number of positive and negative triplets, defined as
\begin{equation}
    \alpha(i, j, k) = \begin{cases}
    \displaystyle\frac{1}{8B}  & \mathrm{if\ } i = j = k \\ 
    \displaystyle-\frac{1}{8(B^3 - B)} 
    & \mathrm{otherwise }
    \end{cases}
\end{equation}

Intuitively, $\mathcal{L}_{\mathrm{inter}}$ is an objective to minimize the area of triangle drawn with positive triplets, thus pulling positive triplets closer in joint embedding space. However, the collapse of entire embedding space into a single point leads to minimization of average positive triplets area mathematically, similar to collapse in Section \ref{3.2}. Referring to Section \ref{4.1} emphasizing the regularization role of intermodal uniformity, we aim to simultaneously shrink positive triangles and expand negative ones.

The intramodal loss is calculated in a encoder-wise manner as intramodal alignment and uniformity are independent to relationships between modalities.
Specifically, TriCL adopts NT-Xent as intramodal losses for each encoder, which is a combination of intramodal alignment and uniformity as shown in equation \ref{eq:nt_xent} \citep{MCLspace, PropCL}. For similarity metric $\mathrm{sim}$, the cosine similarity is used in TriCL.
\begin{align}
    \label{eq:nt_xent}
    \mathcal{L}_{\mathrm{intra}}^{\mathrm{enc}} &= \frac{1}{2B}\sum_{k = 1}^{B} \left(\ell(2k-1, 2k) + \ell(2k, 2k-1)\right)\\
    \text{where}\quad  \ell(i, j) &= \underbrace{-\frac{1}{n \tau}\sum_{i,j}\mathrm{sim}(\bm z^{\mathrm{enc}}_i, \bm z^{\mathrm{enc}}_j)}_{\text{\normalsize intramodal alignment}}
    +  \underbrace{\frac{1}{n}\sum_i\log\sum_{k=1}^{2n} \mathbbm{1}_{k \neq i}\exp(\mathrm{sim}(\bm z^{\mathrm{enc}}_i, \bm z^{\mathrm{enc}}_k)/\tau)}_{{\text{\normalsize intramodal uniformity}}} \nonumber
\end{align}

We unified intramodal alignment loss and intramodal uniformity loss for each individual modality, resulting in the objective function in Equation \ref{eq:decompose} that comprises three intramodal contrastive losses and one intermodal contrastive loss.
$\lambda^{\mathrm{main}}_{\mathrm{intra}}$ and $\lambda_{\mathrm{inter}}$ are hyperparameters controlling weight of the main modality and the intermodal contrastive loss respectively.
\begin{equation}
\label{eq:decompose}
\mathcal{L} = \lambda_{\mathrm{intra}}^{\mathrm{main}}\mathcal{L}_{\mathrm{intra}}^{\mathrm{main}} + \mathcal{L}_{\mathrm{intra}}^{\mathrm{aux1}} + \mathcal{L}_{\mathrm{intra}}^{\mathrm{aux2}} + \lambda_{\mathrm{inter}} \mathcal{L}_{\mathrm{inter}}
\end{equation}

\subsection{More on Embedding Space}
\label{4.3}
We would finish this section by revisiting and continuously discussing the embedding space.
At section \ref{3.2}, we reported a collapse of the embedding space into a `line' while applying NT-Xent loss as an intermodal contrastive loss.
To address this problem, we introduce Triangular Area Loss which is expected to regulate the encoders from falling into the local minima by inspecting the geometry of the embedding space (Section \ref{4.1}).

We believe the last question to be: \textbf{Does Triangular Area Loss solve collapsing problem?}
To give an answer to the question, we would show that 1) Triangle Area Loss mitigates the collapse by dispersing embedding vectors from different modalities, and 2) Joint application of Triangle Area Loss with NT-Xent generates informative embedding spaces through the experiments below.

\paragraph{Triangle Area Loss discriminates modalities by angle.}
Maintaining experimental conditions from Section \ref{3.2}, we apply Triangular Area Loss replacing intermodal NT-Xent loss.
Intramodal alignment and uniformity metrics converge to 0.997 and 1.000 respectively, which indicates the collapse reproduced for each encoder and space falls into a `line'.
Resulting embedding space is also inappropriate for multimodal CL because 0.000 intermodal combined metric implies that the encoder could not distinguish different samples.
However, we found a key to solve collapsing problem from 0.002 of intermodal uniformity, which explains a topology of the space in detail that embedding vectors form several `lines' with diverse angles, rather than a single `line' as in Section \ref{3.2}. 

\paragraph{Joint application of NT-Xent forms `cones'}
Without considering Triangle Area Loss, NT-Xent loss was first applied as both intramodal and intermodal loss.
After careful balancing between a spreading effect of intramodal NT-Xent loss and a collapsing effect of intermodal NT-Xent loss, we could observe an embedding space as a shape of `cone' with 0.660 intramodal alignment.
Yet resulting embedding space had several advantages over the `line' space such as low intramodal uniformity indicating expressive power of the encoder, this joint embedding space is not informative because intermodal combined metric was nearly zero.

\paragraph{Angular diversification of `cones' using Triangle Area Loss}
Hypothesizing that the low information gain from auxiliary modalities stems from the collapsing effect in Section \ref{3.2}, we widen the angles between `cones' using Triangle Area Loss as an intermodal loss.
Consistent to previous observation, angles between `cones' become wider, indicated from decreased intermodal uniformity from 0.091 to 0.079.
Surprisingly, combined metrics of both intramodal and intermodal losses recorded the highest value, which implies that the main encoder can better capture both features available from the main encoder itself and from auxiliary modalities.

\section{Experiments}
\label{experiments}

To assess the performance of TriCL framework and triangle area loss, we implemented TriCL comprising widely-used architectures for each representation: transformer, GNN, and CNN.
We also adopted and implemented three types of augmentation strategies appropriate for strings, graphs, and conformers.
Under these condition, TriCL achieved a state-of-the-art performance on molecular property classification tasks. 
We would present experimental details and explanations in the following.

\subsection{Task Definition}
\paragraph{Datasets}
We pre-trained TriCL on the molecular conformation dataset then fine-tuned on the molecular property prediction downstream tasks. 
As in \citet{graphmvp}, 50k qualified molecules randomly chosen from GEOM dataset \citep{geom} were used for pre-training.
The pre-trained model is fine-tuned and assessed on 8 binary molecular property classification tasks from MoleculeNet \citep{wu2018moleculenet}. Note that since MoleculeNet dataset only contains graph level representation, 3D conformer information is unavailable in downstream tasks. Further details on datasets used are described in Appendix C.

\paragraph{Baselines}
We compared our results with models from well-acknowledged, peer-reviewed works dealing with graph SSL: EdgePred \citep{EdgePred}, InfoGraph \citep{EdgePred}, GPT-GNN \citep{GPT-GNN}, ContextPred \citep{ContextPred}, GraphLoG \citep{GraphLoG}, G-Motif \citep{G-CM}, GraphCL \citep{GraphCL}, and JOAO \citep{JOAO}. 
We also compared our model with 3D structure-aware graph SSL model GraphMVP\citep{graphmvp}.

\subsection{TriCL Implementation}
\paragraph{Encoders} 
Referring to previous graph-based self-supervised learning models \citep{graphmvp, ContextPred, GraphCL, JOAO}, we adopted five layers of Graph Isomorphism Network (GIN) \citep{GIN} as the main encoder. 
For two auxiliary encoders, we used 6 layers of transformers and 4 layers of 3DCNN, which are best suitable for learning representations from 1D SELFIES strings and 3D structures, respectively.
Specifically, the transformer layer is directly adopted from PyTorch and 3DCNN architecture refers to \citet{atom3d}.
The resulting 3 embedding vectors were fed into three consecutive multilayer perceptron layers producing three joint embedding vectors.
\paragraph{Initial Representation}
Starting from the 1D SMILES string randomly chosen from GEOM, we first build the 1D SELFIES string by following \citet{SELFIES}.
We utilized SELFIES rather than SMILES, because SELFIES string representation maintains valid over any types of permutations and mutations thereby much appropriate for reasonable augmentation.
2D graph representations were obtained utilizing RDKit \citep{rdkit} in combination with the PyTorch Geometric.
Atomic coordinates in 3D conformer structures were calculated using RDKit, then voxelized following \citet{atom3d}.
\paragraph{Augmentation} 
We adopted and implemented node drop (ND), node/edge masking (NM), and subgraph masking (SM) augmentation for each representation type and corresponding architectures.
Specifically, we referred \citet{GraphCL} and \citet{MolCLR} for graph augmentations while string and structure augmentation are newly devised.
As described in Section \ref{4.2}, one query representation is augmented twice under the same policy, resulting in two augmented representations for each modality.
The final model used ND for GNN; NM for transformers and CNN; SM for all three architectures.
\paragraph{Optimization}
Triangle Area Loss in combination with NT-Xent loss is applied during pre-training.
Model parameters were optimized using the Adam optimizer, and hyperparameters were tuned through the grid search.

Further details on model implementation are described in Appendix D.

\begin{table}[t]
    \scriptsize
  \caption{Results on the molecular property prediction classification tasks. We report an average test AUC-ROC on 8 downstream tasks with standard deviation inside the parenthesis. Top 1 AUC-ROC score for each task is underlined and bolded. Datasets were scaffold splitted. Baseline performances were adopted from \citet{graphmvp}. Finetuning was repeated under 3 independent seeds $\{0, 1, 42\}$. We report the test AUC-ROC at the epoch which validation AUC-ROC was the highest.}
  \label{cls_table}
  \centering
  \begin{tabular}{llllllllll}
    \toprule
    Pre-training& BBBP          & Tox21         &ToxCast        & SIDER         & ClinTox       & MUV           & HIV           & BACE          & AVG\\
    \midrule
    -           & 65.4(2.4)     & 74.9(0.8)     & 61.6(1.2)     & 58.0(2.4)     & 58.8(5.5)     & 71.0(2.5)     & 75.3(0.5)     & 72.6(4.9)     & 67.21     \\
    \midrule
    EdgePred    & 64.5(3.1)     & 74.5(0.4)     & 60.8(0.5)     & 56.7(0.1)     & 55.8(6.2)     & 73.3(1.6)     & 75.1(0.8)     & 64.6(4.7)     & 65.64     \\
    AttrMask    & 70.2(0.5)     & 74.2(0.8)     & 62.5(0.4)     & 60.4(0.6)     & 68.6(9.6)     & 73.9(1.3)     & 74.3(1.3)     & 77.2(1.4)     & 70.16     \\
    GPT-GNN     & 64.5(1.1)     & 74.2(0.8)     & 62.5(0.4)     & 60.4(0.6)     & 68.6(9.6)     & 73.9(1.3)     & 74.3(1.3)     & 77.2(1.4)     & 68.27.    \\
    InfoGraph   & 69.2(0.8)     & 73.0(0.7)     & 62.0(0.3)     & 59.2(0.2)     & 75.1(5.0)     & 74.0(1.5)     & 74.5(1.8)     & 73.9(2.5)     & 70.10     \\
    ContextPred & 71.2(0.9)     & 73.3(0.5)     & 62.8(0.3)     & 59.3(1.4)     & 73.7(4.0)     & 72.5(2.2)     & 75.8(1.1)     & 78.6(1.4)     & 70.89     \\
    GraphLoG    & 67.8(1.7)     & 73.0(0.3)     & 62.2(0.4)     & 57.4(2.3)     & 62.0(1.8)     & 73.1(1.7)     & 73.4(0.6)     & 78.8(0.7)     & 68.47    \\
    G-Motif     & 66.4(3.4)     & 73.2(0.8)     & 62.6(0.5)     & 60.6(1.1)     & 77.8(2.0)     & 73.3(2.0)     & 73.8(1.4)     & 73.4(4.0)     & 70.14    \\
    GraphCL     & 67.5(3.3)     & 75.0(0.3)     & 62.8(0.2)     & 60.1(1.3)     & 78.9(4.2)     & 77.1(1.0)     & 75.0(0.4)     & 68.7(7.8)     & 70.64   \\
    JOAO        & 66.0(0.6)     & 74.4(0.7)     & 62.7(0.6)     & 60.7(1.0)     & 66.3(3.9)     & 77.0(2.2)     & 76.6(0.5)     & 72.9(2.0)     & 69.57   \\
    \midrule
    GraphMVP-G  & 70.8(0.5)      &\underline{\textbf{75.9(0.5)}} & 63.1(0.2)     & 60.2(1.1)     & 79.1(2.8).   & \underline{\textbf{77.7(0.6)}}     & 76.0(0.1). & 79.3(1.5)  & 72.76   \\
    GraphMVP-C  & \underline{\textbf{72.4(1.6)}} & 74.4(0.2)     & 63.1(0.4)     & \underline{\textbf{63.9(1.2)}}      & 77.5(4.2)     & 75.0(1.0)     & 77.0(1.2)     &81.2(0.9) & 73.07   \\
    \midrule
    \bf{TriCL(OURS)} & \underline{\textbf{72.4(0.4)}} &75.5(0.3) &\underline{\textbf{63.9(0.4)}} &62.0(1.0) &\underline{\textbf{85.4(1.9)}} &77.0(0.8) &\underline{\textbf{78.9(0.5)}} &\underline{\textbf{82.5(1.2)}} &\underline{\textbf{74.71}}\\

    \bottomrule
  \end{tabular}
\end{table}

\subsection{Results on the molecular property classification tasks.}
Results are summarized in Table \ref{cls_table}. TriCL achieved an outstanding average 74.71 AUC on 8 molecular property classification tasks, with best performance on 5 tasks and Top 2 accuracy on 2 tasks.

\begin{table}[t]
\label{ablation_loss}
  \centering
  \subfloat[][{Table 3a: Ablation on objective function. }]{
  \label{sub1}
  \begin{tabular}{l|l|ll}
    \toprule
     \multicolumn{1}{c|}{Intra loss}  & \multicolumn{1}{c|}{Inter loss}  & \multicolumn{1}{c}{Performance}\\
    \midrule
    NT-Xent           & -                 & 72.29  \\
    \midrule
    -                 & NT-Xent           & 71.88  \\ %
    -                 & Triplet Margin    & 71.10  \\ 
    -                 & Triangle Area     & 71.20  \\ %
    \midrule
    NT-Xent           & NT-Xent           & 73.31 \\ %
    NT-Xent           & Triplet Margin    & 73.51 \\ %
    NT-Xent           & Triangle Area     & \textbf{74.71}\\ %
    \bottomrule
  \end{tabular}
  } \quad
  \subfloat[][{Table 3b: Ablation on augmentation methods. }]{
  \label{sub2}
  \begin{tabular}{llll}
    \toprule
    \multicolumn{3}{c}{Augmentation Method}  & \multicolumn{1}{c}{Performance}\\
    \cmidrule(r){1-3} \cmidrule(r){4-4}
    NM & ND & SM   & AUC  \\
    \midrule
    \checkmark & -          & -          & 71.40  \\ %
    -          & \checkmark & -          & 72.38  \\ %
    -          & -          & \checkmark & 71.43  \\ %
    \midrule
    \checkmark & -          & \checkmark & 71.63  \\ %
    -          & \checkmark & \checkmark & \textbf{74.71}  \\ %
    \checkmark & \checkmark & \checkmark & 72.13  \\ %
    
    \bottomrule
  \end{tabular}
  }
    
\end{table}

\subsection{Ablation Study}
\label{5.4}
We then assessed the effects of core components in TriCL by systematically ablating each component while maintaining other settings.
Experiments were performed with three seeds, then an average performance among 8 fine-tuning tasks is reported.
Detailed results are provided in Appendix E.

\paragraph{Effects of the objective function}
To demonstrate the effectiveness of Triangular Area Loss, we conducted an experiment where we fine-tuned the same model with different loss functions and assessed their performance on the MoleculeNet dataset. 
The results in Table \ref{sub1} empirically show that: (1) Pre-training with only intermodal loss performs worse than using only intramodal loss, which is as expected in Section \ref{3.2}. As we discussed, the collapse of the embedding space impedes the main encoder to learn from auxiliary modalities. (2) Joint application of intra and intermodal loss better captures multifaceted features of the sample. As we expected in Section \ref{4.3}, careful application of both losses clearly enhances the performance, compared with intramodal loss. (3) Alignment and uniformity in the embedding space by Triangle Area Loss are beneficial for downstream tasks. As discussed previously, careful design of intermodal loss could encourage the encoder to capture the innate geometry of the embedding space, resulting in the highest performance.

\paragraph{Effects of the augmentation strategy}
We also tested dependency on augmentations by assessing performances after applying various combinations of augmentation strategies during pre-training.
Results in Table \ref{sub2} indicate that the performance highly depends on augmentation strategies.
This dependency is expected since TriCL lacks a metric of similarity between samples without augmentation and only regards representations from the same sample as similar pairs. 
Therefore the only way to learn innate similarities between samples is when two samples generate the same augmented representations.
This gives an important insight that while TriCL conceptually could be implemented for other tasks such as video learning, the quality of an augmentation strategy would be crucial to downstream performance. 
We finally note that augmentation strategies should be carefully curated, as in Table \ref{sub2} applying all available augmentations might actually harm the performance. \citet{GoodView} also discusses this phenomenon and gives an explanation by considering the amount of preserved task-relevant information during augmentation. 

\section{Conclusion}
\label{conclusion}

In this paper, we start by inspecting how alignment and uniformity in intramodal and intermodal contrastive loss construct the joint embedding space. 
To mitigate a line-collapsing problem, we proposed Triangle Area Loss, a novel intermodal contrastive loss that can learn the geometry in terms of angle through contrasting the area of positive triplets and negative triplets.
Under TriCL, a universal trimodal contrastive learning framework, we formulized Triangle Area Loss and discussed the advantages of the embedding space in multimodal representation learning.
Our experimental results demonstrate the outperformance of TriCL compared to existing methods even when only unimodal information is available on downstream tasks.

\paragraph{Generalization and expansion of TriCL}
To the best of our knowledge, learning molecular representation is one of the most natural and general task for multimodal contrastive learning, since (1) instead of arbitrarily chosen data formats, our use of strings, graphs, and conformer structures respectively represent 1D, 2D, and 3D information of molecules, (2) chosen main (GIN) and auxiliary (Transformer and CNN) encoders are the most representative architecture for treating their respective data formats, and (3) our experiment considers the case where only unimodal information is available for downstream tasks, by choosing GIN as the main encoder. Adapting TriCL on tasks where multiple encoders are of the same architecture (such as machine translation) or representations are shared among samples (image tagging) together with analyzing geometric properties of embedding space learned for such tasks is an interesting future work.

\paragraph{Theoretical considerations}
We characterized certain desirable properties of embedding space in terms of alignment and uniformity, and designed TriCL to suffice such objectives. Admittedly, our arguments primarily rely on empirical results and observations. Mathematically rigorous discussion on the topic of joint embedding space learned via tri and even higher modalities remains a difficult and important work, which we strongly believe TriCL would be the great starting point.
    
\pagebreak

{
\bibliographystyle{plainnat}
\bibliography{ref}
}

\newpage

\appendix

\addcontentsline{toc}{section}{Appendix} %
\begin{center}
\part{\centering Appendix} %
\end{center}
\parttoc %
\renewcommand{\contentsname}{Table of contents}
\renewcommand{\baselinestretch}{1.0}\normalsize
\setcounter{table}{0}
\setcounter{figure}{0}
\renewcommand{\thetable}{A\arabic{table}}
\renewcommand{\thefigure}{A\arabic{figure}}

\newpage

\section{Additional Results}
The following section would describe results obtained from additional experiments that were not described in the main manuscript. This would include: (1) Scaled-up pre-training on a large dataset, (2) SELFIES to SMILES alternation test, (3) Performance comparison to the bimodal system. We hope this section would provide a detailed explanation supporting TriCL and Triangular Area Loss, while also deliver useful insights for future research.

\subsection{Pre-training on Large Dataset}
To assess the effects of unlabeled dataset size for pre-training, we pre-train TriCL with larger numbers of molecules in GEOM. 
As shown in Table \ref{scale_up}, increased size of dataset could be either beneficial or not depending on tasks. 
In case of Tox21, ToxCast and MUV, AUC-ROCs are gradually increased while for BBBP, ClinTox, HIV and BACE, AUC-ROCs are higher when pre-trained with smaller dataset. We could conclude that \textit{unlabeled dataset should also be curated task-specifically} to achieve optimal performance, because the model could be `distracted' by irrelevant samples. 

\begin{table}[h]
    \scriptsize
  \caption{Downstream performances of TriCL pre-trained on dataset with different size. TriCL was identically pre-trained and fine-tuned except for the pre-training dataset size.}
  \label{scale_up}
  \centering
  \begin{tabular}{l|llllllll|l}
    \toprule
    \# Compounds & BBBP & Tox21         &ToxCast        & SIDER         & ClinTox       & MUV           & HIV           & BACE          & AVG \\
    \midrule
    50k         & \underline{\textbf{72.4(0.4)}} &75.5(0.3) &63.9(0.4) &62.0(1.0) &\underline{\textbf{85.4(1.9)}} &77.0(0.8) &\underline{\textbf{78.9(0.5)}} &\underline{\textbf{82.5(1.2)}} &74.71 \\ 
    100k        & 71.7(0.5) & 75.7(0.5) & 64.1(0.1) & \underline{\textbf{62.1(0.6)}} & 81.2(2.2) & 77.6(0.8) & 78.6(0.2) & 82.3(0.8) & 74.16 \\ 
    200k        & 72.1(0.8) & \underline{\textbf{76.3(0.3)}} & \underline{\textbf{64.7(0.3)}} & 61.6(0.3) & 85.2(1.0) & \underline{\textbf{78.3(1.4)}} & 78.3(0.8) & 82.1(1.1) & \underline{\textbf{74.83}}\\
    \bottomrule
  \end{tabular}
\end{table}

\subsection{Pre-training with SMILES String Representation}
Instead of using the most common molecular text representation SMILES, TriCL adopts SELFIES string as a 1D representation of the molecule.
We believe that SELFIES is a better representation for extracting chemically significant properties because: (1) SELFIES keep representing valid molecules even when any types of mutations are applied on the original string. Thus, using SELFIES could help the model to learn from chemically valid augmented representations, which obviously benefit learning more meaningful relationships between augmentations than using SMILES. (2) In SELFIES string, all tokens possess semantic meanings, while in SMILES some tokens play only syntactic roles. Since each token represents a specific chemical unit in SELFIES, the utilization of node/edge masking and subgraph masking techniques generate meaningful augmented representations. 
\setlength{\tabcolsep}{3.5pt}
\begin{table}[h]
    \scriptsize
  \caption{Performance of TriCLs applying SMILES/SELFIES(OURS) as 1D auxiliary representation.\vspace{0.1cm}}
  \label{smiles}
  \centering
  \begin{tabular}{lll|llllllll|c|c|c}
    \toprule
    NM & ND & SM          & BBBP & Tox21         &ToxCast        & SIDER         & ClinTox       & MUV           & HIV           & BACE          & AVG & OURS & $\Delta$ \\
    \midrule
    \checkmark & - & -                      & 66.0(3.0) & 75.5(0.4) & 64.2(0.4) & 61.1(1.6) & 64.8(5.4) & 76.7(0.2) & 77.2(0.2) & 77.8(1.1) & 70.51 & \textbf{71.40} & \textbf{+0.89}\\
    - & \checkmark & -                      & 69.5(0.8) & 74.1(0.5) & 62.7(1.1) & 61.6(0.5) & 76.8(0.8) & 74.3(2.4) & 75.4(2.0) & 81.0(1.7) & 71.94 & \textbf{72.38} & \textbf{+0.44}\\
    - & - & \checkmark                      & 68.7(0.4) & 74.6(0.1) & 63.1(0.8) & 60.0(1.0) & 76.5(2.7) & 73.5(1.2) & 75.2(0.8) & 78.4(3.2) & 71.25 & \textbf{71.43} & \textbf{+0.18}\\
    \midrule
    \checkmark & - & \checkmark             & 67.6(4.0) & 75.0(1.1) & 62.2(0.7) & 59.7(1.0) & 70.6(2.2) & 77.0(0.6) & 76.6(0.3) & 77.5(2.3) & 70.77 & \textbf{71.63} & \textbf{+0.86}\\
    - & \checkmark & \checkmark             & 71.8(1.0) & 74.5(0.7) & 63.5(0.2) & 60.7(0.9) & 79.1(2.4) & 76.0(2.0) & 76.6(1.2) & 81.6(2.0) & 72.98  & \textbf{74.71} & \textbf{+1.73}\\
    \checkmark & \checkmark & \checkmark    & 70.8(0.8) & 73.2(0.4) & 61.4(0.2) & 60.5(2.0) & 65.4(4.7) & 74.1(1.7) & 75.2(1.6) & 71.5(1.7) & 69.01 & \textbf{72.13} & \textbf{+3.12}\\
    \bottomrule
  \end{tabular}
\end{table}

\subsection{Comparison to Bimodal CL}
We validate the necessity of each auxiliary modality by implementing a bimodal CL framework using NT-Xent intermodal loss and measuring performances of models on the same downstream tasks. 
As stated in Table \ref{bimodal}, GNNs trained with two auxiliary modalities shows better performances than GNNs trained with one auxiliary modality. 
This result might mislead that TriCL's performance stems from additional modalities.
However, the facts that TriCL showed significantly better performance than the pairwise trimodal contrastive learning method implies that \textbf{\textit{TriCL takes advantage of Triangular Area Loss, which better distill diverse perspectives of auxiliary modalities to the main encoder by contrasting triplets and considering the geometry of them simultaneously}}.

\setlength{\tabcolsep}{5pt}
\begin{table}[H]
    \scriptsize
  \caption{Comparison to Bimodal CL. All other settings remained the same.}
  \label{bimodal}
  \centering
  \begin{tabular}{ll|llllllll|l}
    \toprule
    Encoder & Loss                & BBBP & Tox21 &ToxCast & SIDER & ClinTox  & MUV  & HIV  & BACE & AVG \\
    \midrule
    1D+2D & NT-Xent               & 71.1(0.7) & 75.0(0.5) & 64.0(0.8) & 61.0(0.6) & 79.9(2.2) & 76.3(0.6) & 76.8(0.8) & 80.1(1.8) & 73.02 \\ 
    3D+2D & NT-Xent               & 72.2(1.8) & 74.8(1.0) & 64.3(0.4) & 58.7(1.4) & 78.2(3.4) & 78.1(1.2) & 77.1(1.1) & 78.9(3.2) & 72.80 \\
    \midrule
    1D+2D+3D & NT-Xent            & 71.1(0.3) & 75.0(0.6) & 63.6(0.5) & 60.6(1.2) & 81.6(4.4) & 76.7(1.0) & 77.4(0.6) & 80.4(1.5) & 73.31 \\ 
    1D+2D+3D & Triangular         & 72.4(0.4) & 75.5(0.3) & 63.9(0.4) & 62.0(1.0) & 85.4(1.9) & 77.0(0.8) & 78.9(0.5) & 82.5(1.2) & \underline{\textbf{74.71}} \\
    \bottomrule
  \end{tabular}
\end{table}
\section{Case Study}
\label{casestudy}
We would support our experiments about an embedding space and performances of TriCL by assessing TriCL on widely-used, domain-acknowledged independent validation dataset. 

\subsection{DUD-E dataset and GPCR target proteins}
\textit{DUD-E} (Database of Useful Decoys: Enhanced) \citepApp{DUDE} provides challenging negative samples (`decoys') for the protein-molecule docking task. 22,886 active compounds and their affinities against 102 protein targets are contained. 
\begin{itemize}
    \item For each positive (`active') protein-binding compound, 50 decoy molecules having \textbf{\textit{similar physico-chemical properties}} but \textbf{\textit{dissimilar in 2D graph topologies}} are also involved.
    \item Therefore \textbf{\textit{we expect pre-trained GNN to map active compounds and corresponding decoys near in space while discriminating irrelevant molecules}}, so that GNN can concentrate on capturing sophisticated differences in graph during the fine-tuning phase.
    \item This could be measured by applying the same metrics with Section \ref{3.2}, alignment and uniformity; \textbf{\textit{alignment between molecules targeting the same protein and uniformity between irrelevant molecules should be high}} for better performance.
\end{itemize}

Among subsets of DUD-E protein targets, we focused on \textit{GPCR}(G protein-coupled receptor)-binding active compounds and corresponding decoys because of three reasons: 
\begin{enumerate}
    \item GPCRs are involved in multiple different signaling pathways crucial in life of the cell which makes GPCRs important protein targets \citepApp{gpcr4}.
    \item GPCRs are highly dynamic entailing huge conformational change during activation, which necessitates the delicate 3D design of drug structures \citepApp{gpcr1, gpcr2}.
    \item Small molecules without careful structural design can bind multiple substructures of GPCRs which can cause severe side effects \citepApp{gpcr3}. 
\end{enumerate}

DUD-E GPCR subset comprises five specific GPCR targets - AA2AR (Anti-adenosine A2A receptor), ADRB1 (Adrenoceptor beta 1), ADRB2 (Adrenoceptor beta 2), CXCR4 (C-X-C chemokine receptor type 4), and DRD3 (Dopamine receptor D3). We first verified the embedding space made by all GPCRs, then analyzed the embedding spaces of molecules targeting each specific target.

\subsection{Embedding Space Properties}
Using pre-trained TriCL, we assessed the alignment of generated embedding space made from active compounds of GPCRs. Uniformity was measured by measuring cosine similarities with respect to irrelevant compounds which bind to other proteins. Combined metric was calculated by $\mathrm{(Align - \abs{Uniform})}$, as defined from Table \ref{alignment_metric}. For instance-wise analysis, 500 active compounds (for AA2AR, ADRB1, ADRB2 and DRD3) and 200 active compounds (for CXCR4) were randomly selected from each target with corresponding decoys then average cosine similaries are reported. We compared embedding space metrics with the base GNN encoder trained using only NT-Xent loss as an intramodal contrastive loss. 

\begin{table}[h]
  \caption{Case study on GPCR-binding compounds. Alignment metric is the average cosine similarity between all active compounds targeting GPCRs or the same GPCR (higher is better). Uniformity metric is the average cosine similarity between GPCR-targeting compounds and others (close to 0 is better). Combined metric refers to $\mathrm{(Align - \abs{Uniform})}$ (higher is better).}
  \label{GPCR}
  \centering
  \small  
  \begin{tabular}{c|rrr|rrrrr}
    
    \toprule
    \multicolumn{1}{c|}{\multirow{2}{*}{}} & \multicolumn{3}{c|}{GPCR active compounds} & \multicolumn{5}{c}{Target Instances (Alignment)}
    \\\cmidrule(r){2-9}
    & Align & Uniform & Combined              & AA2AR & ADRB1 & ADRB2 & CXCR4 & DRD3 \\ 
    \midrule
    GNN (Unimodal CL)         & 0.574 & 0.546 & 0.028  & \textbf{0.317} & 0.324 & 0.324 & 0.233 & 0.388\\ 
    TriCL       & \textbf{0.602} & \textbf{0.316} & \textbf{0.286}  & 0.299 & \textbf{0.368} & \textbf{0.384} & \textbf{0.381} & \textbf{0.458} \\
    \bottomrule
  \end{tabular}
\end{table}

As shown in Table \ref{GPCR}, TriCL maps GPCR-binding active compounds near in space (high alignment) while discriminating others (low uniformity). This implies that TriCL indeed captures additional information from auxiliary modularities and these information helps discriminating molecular targets in specific real-world problems. Furthermore, TriCL also outperformed in aligning compounds and decoys targeting the same GPCR target instance. 

\subsection{Active compounds and Decoys Examples}
Here, we provide several examples of a set of active compounds and corresponding decoys. 
Decoy molecules are structurally similar to corresponding acitve compound, but their 2D graph representations have huge differences. 
TriCL can recognize structural similarities between active and decoys by simultaneously contrasts with auxiliary modalities. 

\begin{figure}[h]
  \centering
    \includegraphics[width=\linewidth]{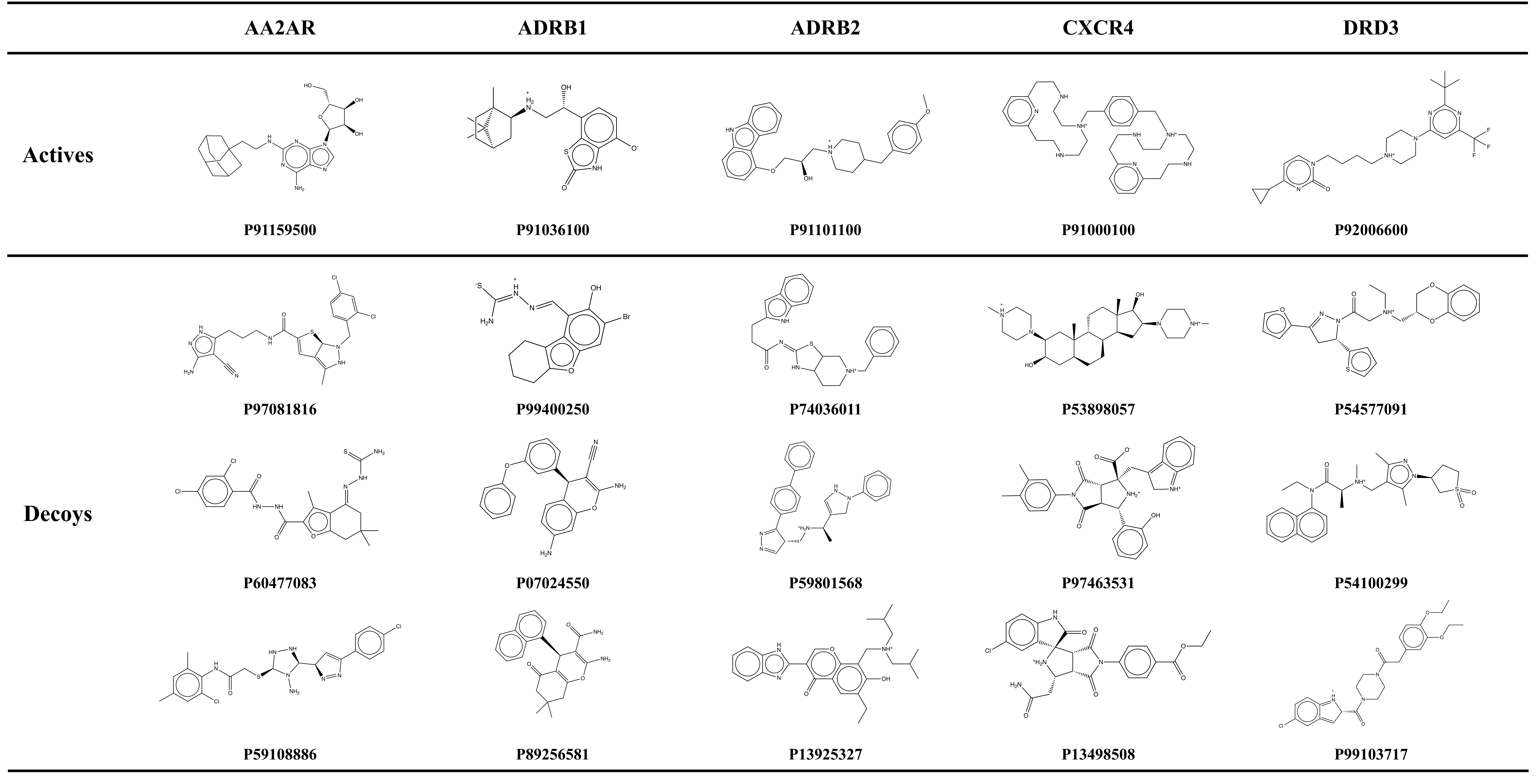}
  \caption{Selected active and decoy compounds in DUD-E GPCR subset. Labels below each structures are protonation codes, provided in DUD-E dataset.}
  \label{fig:fig1}
\end{figure}
\section{Dataset and Baseline Models Overview}
\label{App_data}

\subsection{Pre-training Dataset Overview}
\paragraph{GEOM} Geometric Ensemble Of Molecules (GEOM) is a dataset of high-quality conformers for 317,928 mid-sized organic molecules with experimental data \citepApp{geom_appendix}. Conformers in GEOM are generated with the CREST program \citepApp{CREST} which can generate reliable and accurate structures by using extensive sampling based on the semi-empirical extended tight-binding method (GFN2-xTB) \citepApp{GFN2-xTB}. CC BY 4.0. 50k molecules and corresponding conformer were selected and utilized for pre-training TriCL. Results using full data ($\sim$200k) can be found in Table \ref{scale_up}.

\subsection{Fine-tuning Dataset Overview}
For downstream finetuning-tasks, we used MoleculeNet \citepApp{moleculenet_appendix}. All data were split by scaffold. Most datasets have no license specification; we mark them as MIT License based on deepchem \citepApp{deepchem}

\textit{BBBP} is a binary classification task that aims to predict the ability of a drug whether it penetrates the blood-brain barrier(BBB), a membrane separating circulating blood and brain extracellular fluid. Since BBB penetrated drugs might directly affect the central nervous system, BBBP is a crucial challenge in drug development. \citepApp{BBBP_sub}. This dataset curated by \citetApp{BBBP} contains 1,975 drugs. MIT License.

\textit{Tox21} is a multitask classification dataset that was curated from the "Toxicology in the 21st Century" initiative and has been used in the 2014 Tox21 Data Challenge \citepApp{Tox21}. Tox21 comprises qualitative toxicity measurements of 7,831 compounds on 12 different targets including nuclear receptors and stress response pathways. Nearly 6,000 drugs are annotated for each target. MIT License.

\textit{ToxCast} is a similar data collection as Tox21, but much intensive toxicity dataset including qualitative results of 617 \textit{in vitro} high-throughput screening assays on 8,575 compounds \citepApp{ToxCast}. Depending on a specific task, hundreds to thousands of drugs are labeled. MIT License.

\textit{ClinTox} is a task discriminating FDA-approved drugs from rejected drugs that have failed at the clinical trial stage, for the reason of toxicity \citepApp{ClinTox}. ClinTox is composed of two binary classification tasks dealing with: (1) clinical trial toxicity and (2) FDA approval status. A total of 1,478 compounds are curated in the ClinTox dataset. MIT License.

\textit{SIDER} (Side Effect Resource) is about marketed drugs and adverse drug reactions (ADR) \citepApp{SIDER}. The raw SIDER dataset is annotated hierarchically; we used the common version which grouped the side effects of 1,427 drugs into 27 system organ classes. CC BY 4.0.

\textit{MUV} (Maximum Unbiased Validation) is a virtual screening benchmark dataset curated by using the refined nearest neighbor analysis on PubChem bioactivity data \citepApp{MUV}. MUV dataset contains 93,087 compounds for 17 subtasks from pairs of primary high-throughput screening assays and confirmatory dose-response experiments. MIT License.

\textit{HIV} was annotated by the Drug Therapeutics Program (DTP) AIDS Antiviral Screen \citepApp{HIV_appendix}. The raw dataset classified drugs into three categories - confirmed inactive (CI), confirmed active (CA), and confirmed moderately active (CM). This data was re-curated into binary classification between inactive (CI) and active (CA and CM). A total of 41,127 compounds are contained. CC BY 4.0.

\textit{BACE} deals with qualitative binding results for human beta-secretase 1 (BACE-1) \citepApp{BACE}. A raw BACE dataset containing quantitative IC50 of 1,513 compounds reported in the scientific literature was binarized for building a classification task. MIT License.

\subsection{Baseline Models Overview}

\paragraph{Graph-only SSL Baseline}
\textit{EdgePred} \citetApp{EdgePred_appendix} masked node/edge attributes, then pre-trained GNN to obtain the corresponding node/edge embeddings. \textit{InfoGraph} \citetApp{InfoGraph_appendix} adopted student-teacher model by maximizing the mutual information between the graph-level representation of InfoGraph and substructure representations of existing supervised methods. \textit{GPT-GNN} \citetApp{GPT-GNN_appendix} introduced a self-supervised attributed graph attribute and edge generation so that the model learns capturing inherent dependencies between node attributes and graph structures. \textit{ContextPred} \citetApp{ContextPred_appendix} contrasts node features with subgraph structures by pre-training GNN to map nodes in similar structural contexts to nearby embeddings. \textit{GraphLoG} \citetApp{GraphLoG_appendix} learned global semantic structure by introducing hierarchical prototypes and maximizing the data likelihood with respect to GNN parameters via an online expectation-maximization (EM) algorithm. \textit{GraphCL} \citetApp{GraphCL_appendix} introduced four types of augmentation - node drop, edge perturbation, attribute masking, and subgraph sampling - and performed task-wise analysis.
\textit{JOAO} \citetApp{JOAO_appendix} adopted a plug-and-play framework for optimizing a joint augmentation method over four augmentation strategies.
\paragraph{Structure-aware SSL Baseline}
\textit{GraphMVP} \citetApp{graphmvp_appendix} incorporated 3D structural view into GNN by maximizing mutual information between 3D graph representation and 2D graph representation using both contrastive and generative SSL methods.\\
\newpage
\section{TriCL Implementation Details}
In this section, we would explain details in implementation of TriCL. 
\label{App_Implement}

\subsection{TriCL framework overview}
As briefly described in Section \ref{4.2}, TriCL is devised to pre-train a \texttt{main} encoder while contrasting with two other auxiliary modalities, \texttt{aux1} and \texttt{aux2}. Before specifying architectural details of submodules in TriCL, we would first highlight key properties of TriCL:

\begin{enumerate}
    \item \textbf{Modularity} The main advantage of TriCL is its modularity: three networks could be independently designed and transplanted into the TriCL framework. Following Triangular Area Loss would force the \texttt{main} encoder to learn similarities and differences between inputs while borrowing diverse viewpoints from auxiliary modalities.
    \item \textbf{Universality} TriCL is indeed suitable for any type of trimodal pre-training tasks by devising appropriate encoders which can effectively extract useful properties of the input having diverse formats. For any data format that comprises three distinct components, TriCL can provide a universal method for pre-training one \texttt{main} encoder representing the input object.
    \item \textbf{Unity} TriCL learns relationships among three modalities simultaneously, not by learning pairwise relationships between two modalities. Triangular Area Loss is fundamentally designed to consider geometry made from three representations simultaneously by utilizing the area of the triangle. This property encourages the model to learn much-balanced embedding space affected by three networks.
\end{enumerate}

In the following sections, details for implementing TriCL showing three key properties are described.

\subsection{TriCL Algorithm}
The formal definition of TriCL is provided in Algorithm \ref{alg:TriCL}. For detailed explanations including the definition of NTXent loss and $\alpha, \mathrm{Area}$, please refer to Section \ref{4.2}.

\begin{algorithm}[H] \caption{TriCL main learning algorithm}
\label{alg:TriCL}
\setstretch{1.25}
\begin{algorithmic}
\STATE \textbf{input:} batch size $B$, encoder model $\mathrm{main},\mathrm{aux_1},\mathrm{aux_2}$, molecule representation $\bm m$

\FOR{sampled minibatch $\{\bm m^{\mathrm{model}}_i\}_{1}^B$}
    \FOR{$\mathrm{model}\in \{\mathrm{main, aux_1, aux_2}\}$}
    \FOR{$k \in \{1, \ldots, B\}$}
    \STATE{generate augmented input $\bx^\mathrm{model}_{2k-1}, \bx^\mathrm{model}_{2k}$ from 
    $\bm m^\mathrm{model}_k$}
    \STATE{compute $\bz^\mathrm{model}_{2k-1}, \bz^\mathrm{model}_{2k} = \mathrm{model}(\bx^\mathrm{model}_{2k-1}), \mathrm{model}(\bx^\mathrm{model}_{2k})$}
    \ENDFOR
    \STATE{$\mathcal{L}_\mathrm{model}$ = $\mathcal{L}_\mathrm{model} + \mathrm{NTXent}(\bz^\mathrm{model})$}
    \ENDFOR 
    \STATE{$\mathcal{L}_{\mathrm{inter}} = 0$}
    \FOR{$i, j, k \in \{1, \ldots, 2B\}$}
    \STATE{$\mathcal{L}_{\mathrm{inter}} = \mathcal{L}_{\mathrm{inter}} + \alpha\big(\ceil{i/2}, \ceil{j/2}, \ceil{k/2}\big) \cdot \mathrm{Area}(\bz^1_{i}, \bz^2_{j}, \bz^3_{k})^2$}
    \ENDFOR
    \STATE{$\mathcal{L} = \lambda_{\mathrm{main}}\mathcal{L}_\mathrm{main} + \mathcal{L}_\mathrm{aux_1} + \mathcal{L}_\mathrm{aux_2} + \lambda_{\mathrm{inter}}\mathcal{L}_\mathrm{inter}$}
    \STATE{update parameters of each model to minimize $\mathcal{L}$}
\ENDFOR
\STATE{\textbf{return} encoder $\mathrm{main}$}
\end{algorithmic}
\end{algorithm}

\subsection{Resources}
TriCL is pre-trained using NVIDIA GeForce RTX 3090 for 2 hours on average per each pre-train.
Pre-trained GNN is fine-tuned using NVIDIA GeForce RTX 2080 Ti; about 1 hour was spent for fine-tuning on 8 downstream tasks.

\subsection{Encoding Details}
\textit{Main Encoder }
GNN architecture was directly adopted from \citetApp{graphmvp_appendix} which is also identical to other baseline models. 5 GIN layers were stacked, with 300 hidden dimension and 0.5 dropout ratio. Node embedding vectors at the last layer was mean-pooled.

\textit{Auxiliary Encoder 1 }
Transformer modules were adopted from PyTorch. 6 transformer encoder was sequentially connected. For self-attention, hidden dimension was 64 and 8 heads used for multihead attention. For feed forward layer, hidden dimension was 64. 0.1 dropout ratio was applied for both layers. Other details not described here is specified in Section \ref{IRD}.

\textit{Auxiliary Encoder 2 }
CNN architecture was directly adopted from \citetApp{atom3d_appendix}, which utilized PyTorch Conv3d layer. Starting from 12 in\_channel dimension, 4 3D convolution layers were applied with 3 convolution kernel size, 1 stride and 0 padding. 3D representation tensor was flattened and reduced via one fully connected layer. Dropout ratio was 0.1 for convolution layer, and 0.25 for fully connected layer.

\subsection{Representation Details}
\label{IRD}
\textit{Main Representation}
For graph representation, we followed \citetApp{ContextPred_appendix}. Node feature comprise one atomic number and four chirality features. Edge feature is composed of bond type and bond direction. This representation method was identical to \citetApp{graphmvp_appendix} and other baseline methods. We provide details of node and edge features in Table \ref{graph_feature}.

\begin{table}[h]
  \caption{Node and edge features used in TriCL.}
  \label{graph_feature}
  \centering
  \begin{tabular}{lll}
    \toprule
    Feature type & Feature name & Range\\
    \midrule
    Node feature & Atomic number & [1, 119] \\
                 & Chirality & {unspecified, tetrahedral CW, tetrahedral CCW, other} \\
    \midrule
    Edge feature & Bond type & {single, double, triple, aromatic} \\
                 & Bond direction & {none, end-upright, end-downright}\\
    \bottomrule
  \end{tabular}
\end{table}

\textit{Auxiliary Representation 1}
For string representation, we followed \citetApp{SELFIES_appendix} to convert SMILES strings to SELFIES strings. 
Then, generated SELFIES string was vectorized following pre-defined tokenization rule. Atomic symbols were converted according to atomic vocabulary, which annotates non-metal 99 common atomic tokens. This common atomic tokens were extracted from 10M PubChem dataset \citepApp{pubchem}. Remaining metal tokens not involved in 99 tokens were represented as one metal token [Me]. This abstraction originates from chemical insights that most drugs do not involve metal elements and most metal elements show similar chemical properties compared to non-metal elements. Other tokens representing branches and rings were thoroughly defined. We provide details of SELFIES tokenization rule in Table \ref{selfies_feature}.

\begin{table}[h]
  \caption{SELFIES tokenization rule in TriCL. This includes augmentation tokens.}
  \label{selfies_feature}
  \centering
  \begin{tabular}{llll}
    \toprule
    Components &  Representation & \# Tokens & Token Number \\
    \midrule
    Padding / No Operation & [NOP]                  & 1     & 0 \\
    Atom-masking           & [MASK\_AT]             & 1     & 1 \\
    Branch/Ring-masking    & [MASK\_BO]             & 1     & 2 \\
    Metals                 & [Me]                   & 1     & 3 \\
    Special Classification & [CLS]                  & 1     & 4 \\
    Common Atoms           & [\#B] - [S]            & 99    & 5-103 \\
    Branches               & [Branch1] - [\#Branch3] & 9     & 104-112 \\
    Rings                  & [Ring1] - [-/Ring3]    & 36    & 113-138 \\
    \bottomrule
  \end{tabular}
\end{table}

\textit{Auxiliary Representation 2}
For structure representation, we used RDKit to generate conformers of molecules.
Predefined 10 common non-metal elements were mapped to corresponding feature integers, and other metal elements were abstracted to a single metal integer. The structure was voxelized by 7.5 \AA \: radius of grids, with 1.0 \AA \: resolution, after random rotation. We provide details of element feature representation in Table \ref{structure_feature}.

\begin{table}[h]
  \caption{Structure featurization in TriCL. This includes a masking integer.}
  \label{structure_feature}
  \centering
  \begin{tabular}{lllllllllllll}
    \toprule
    Integer & 0 & 1 & 2 & 3 & 4 & 5 & 6 & 7 & 8 & 9 & 10 & 11 \\
    \midrule
    Element & Mask & H & C & N & O & F & Cl & Br & P & S & B & Metal \\
    \bottomrule
  \end{tabular}
\end{table}

\newpage

\subsection{Augmentation Details}
We would provide a detailed augmentation strategies for each representations. Specifically, we adopted or implemented node drop (ND), node masking (NM), and subgraph masking (SM). Note that each input molecule is augmented two times in terms of each representation format, generating 6 augmented representations. For mixed augmentation settings, independent augmentation strategies were applied sequentially. Predefined ratio of augmentation for ND, NM, and SM were 0.2, 0.2, and 0.05 respectively.

\textit{Main Graph Augmentation}
For NM and SM, we referred \citetApp{MolCLR_appendix}. Predefined ratio of random node and edge features were masked to zero vector in NM augmentation. For SM augmentation, predefined ratio of node and edge features were masked to zero vector, but each node and edge were selected as adjacent components to the randomly selected anchor node.
For node drop, we referred \citetApp{GraphCL_appendix}. Predefined ratio of random nodes were deleted from the graph; edges connected to the deleting nodes were also deleted.

\textit{Auxiliary String Augmentation}
For NM, predefined ratio of random atom tokens and bond tokens were masked to [MASK\_AT] and [MASK\_BO], respectively. 
For SM, predefined ratio of tokens were masked similarly, but tokens were selected from adjacent tokens from randomly selected anchor atom token. To maintain syntax of SELFIES string, the whole subsequence of branch and ring components were masked simultaneously.
For graph ND setting, NM is applied instead.

\textit{Auxiliary Structure Augmentation}
For NM, predefined ratio of random atoms were masked to `M' element.
For SM, predefined ratio of atoms were masked similarly, but atoms were selected from nearest atoms from the randomly selected hetero atom. 
For graph ND setting, NM is applied instead.

\subsection{Optimization Details}
\textit{Pre-training}
Parameters in three encoders and final readout layer were optimized by using Adam optimizer, under weight decay 1e-5. The learning rate was scheduled via cosine annealing scheduler, with initial learning rate 0.0005 and 10 warm-up epoch setting. TriCL is pre-trained 100 epochs. During the warm-up 5 epochs, intermodal loss was not backpropagated.

\textit{Fine-tuning}
For fine-tuning, we followed \citetApp{graphmvp_appendix} and other baseline models for fair comparison. Parameters were optimized via Adam optimizer, under learning rate 0.001.
\newpage
\section{Detailed Results}
Here, we provide the full results of experiments described in Section \ref{experiments}. Specifically, we first illustrated the whole measured AUC-ROC of three seeds on 8 MoleculeNet tasks. Then, task-wise statistics corresponding to Table \ref{sub1} and Table \ref{sub2} would be described.

\subsection{Full results of Table \ref{cls_table}}
Detailed data measured from TriCL is described below. The pre-trained TriCL is tested three times under three fixed independent seeds. We then reported the average and standard deviation in Table \ref{cls_table}.
\begin{table}[h]
    \footnotesize
  \caption{Detailed Results of Table 2.}
  \label{cls_table_detail}
  \centering
  \begin{tabular}{ll|llllllll|l}
    \toprule
    Model & Seed & BBBP          & Tox21         &ToxCast        & SIDER         & ClinTox       & MUV           & HIV           & BACE          & AVG\\
    \midrule
          & 0  & 72.81 & 75.39 & 64.23 & 60.99 & 85.80 & 78.04 & 78.57 & 83.10 & 74.64\\
    TriCL & 1  & 72.07 & 75.82 & 64.03 & 61.97 & 87.37 & 78.06 & 79.36 & 81.20 & 74.99\\
          & 42 & 72.08 & 75.33 & 63.52 & 63.13 & 83.71 & 77.55 & 78.92 & 83.32 & 74.50\\
    \bottomrule
  \end{tabular}
\end{table}

\subsection{Full results of Table \ref{sub1}}
Identical to the experiment above, experiments were repeated three times under three independent seeds - 0, 1, and 42; we report the average AUC-ROC and standard deviation of them in Table \ref{3(a)_table_detail}. We marked top 1 values with \textbf{\underline{bold and underline}}. 
\begin{table}[h]
    \tiny
  \caption{Detailed Results of Table 3(a).}
  \label{3(a)_table_detail}
  \centering
  \begin{tabular}{ll|llllllll|l}
    \toprule
    Intra loss & Inter loss & BBBP          & Tox21         &ToxCast        & SIDER         & ClinTox       & MUV           & HIV           & BACE          & AVG\\
    \midrule
    NT-Xent & -                 & 70.6(1.9) & 75.1(0.2) & \underline{\textbf{63.9(0.4)}} & 60.1(1.4) & 80.4(3.0) & 74.2(1.8) & 77.1(1.2) & 76.8(1.7) & 72.29\\
    \midrule
    - & NT-Xent                 & 69.7(1.8) & 75.2(0.5) & 63.4(0.5) & 60.7(1.5) & 71.3(4.3) & \underline{\textbf{78.2(1.6)}} & 78.3(4.7) & 78.3(4.7) & 71.88\\
    - & Triplet Margin          & 70.3(1.5) & 74.9(0.8) & 62.3(1.4) & 59.3(1.0) & 70.6(1.0) & 73.9(2.2) & 77.8(0.3) & 79.6(1.8) & 71.10\\
    - & Triangular Area         & 69.1(0.8) & 75.2(0.5) & 62.8(0.8) & 61.9(1.2) & 68.4(3.9) & 76.3(0.1) & 77.5(1.4) & 78.5(6.8) & 71.20\\
    \midrule
    NT-Xent & NT-Xent           & 71.1(0.3) & 75.0(0.6) & 63.6(0.5) & 60.6(1.2) & 81.6(4.4) & 76.7(1.0) & 77.4(0.6) & 80.4(1.5) & 73.31\\
    NT-Xent & Triplet Margin    & 70.9(0.6) & 74.2(0.5) & 64.1(0.4) & 61.1(0.9) & 81.4(3.1) & 75.7(2.2) & 77.7(0.2) & \underline{\textbf{82.9(1.1)}} & 73.51\\
    NT-Xent & Triangular Area   & \underline{\textbf{72.4(0.4)}} &\underline{\textbf{75.5(0.3)}} &\underline{\textbf{63.9(0.4)}} &\underline{\textbf{62.0(1.0)}} &\underline{\textbf{85.4(1.9)}} &77.0(0.8) &\underline{\textbf{78.9(0.5)}} &82.5(1.2) &\underline{\textbf{74.71}}\\
    \bottomrule
  \end{tabular}
\end{table}

\subsection{Full results of Table \ref{sub2}}
Experiments were repeated three times under three independent seeds - 0, 1, and 42; we report the average AUC-ROC and standard deviation of them in Table \ref{3(b)_table_detail}. We marked top 1 values with \textbf{\underline{bold and underline}}. 
\begin{table}[h]
    \scriptsize
  \caption{Detailed Results of Table 3(a).}
  \label{3(b)_table_detail}
  \centering
  \begin{tabular}{lll|llllllll|l}
    \toprule
    NM & ND & SM          & BBBP & Tox21         &ToxCast        & SIDER         & ClinTox       & MUV           & HIV           & BACE          & AVG\\
    \midrule
    \checkmark & - & -                      & 68.8(0.1) & \underline{\textbf{75.9(0.1)}} & 62.8(0.1) & 61.5(1.1) & 69.9(4.5) & 76.8(0.8) & 77.1(0.1) & 78.5(1.5) & 71.40\\
    - & \checkmark & -                      & 70.5(0.8) & 75.1(1.1) & 63.4(0.7) & 59.8(0.6) & 77.9(5.9) & 76.1(0.7) & 77.3(0.8) & 78.8(0.8) & 72.38\\
    - & - & \checkmark                      & 69.6(0.2) & 75.3(0.5) & 63.0(0.7) & 61.1(0.9) & 75.2(6.9) & 75.4(0.8) & 77.6(0.8) & 74.3(6.9) & 71.43\\
    \midrule
    \checkmark & - & \checkmark             & 70.1(3.3) & 75.0(0.2) & 63.2(0.5) & \underline{\textbf{62.1(1.0)}} & 69.7(8.0) & 75.4(1.1) & 77.9(0.8) & 76.9(1.6) & 71.63\\
    - & \checkmark & \checkmark             &\underline{\textbf{72.4(0.4)}} &75.5(0.3) &\underline{\textbf{63.9(0.4)}} &62.0(1.0) &\underline{\textbf{85.4(1.9)}} &\underline{\textbf{77.0(0.8)}} &\underline{\textbf{78.9(0.5)}} &\underline{\textbf{82.5(1.2)}} &\underline{\textbf{74.71}}\\
    \checkmark & \checkmark & \checkmark    & 70.5(2.0) & 74.3(0.8) & 63.2(0.6) & 60.0(0.8) & 79.6(0.6) & 75.4(1.1) & 75.8(1.1) & 78.2(1.8) & 72.13\\
    \bottomrule
  \end{tabular}
\end{table}
\newpage
\section{Limitations, Future Directions and Broader Impacts}
Lastly, we would carefully illustrate limitations of TriCL and introduce future research topics that can directly utilize or inspired by TriCL.  Then we would finish by discussing possible societal impacts.

\subsection{Limitations and Future Directions}
\begin{enumerate}
    \item \textit{Theoretical Considerations} As discussed in Section \ref{conclusion}, our arguments primarily rely on empirical results and observations. Mathematically rigorous discussion on joint embedding spaces and optimization would be an important future challenge.
    
    \item \textit{Application on Other Tasks} Also discussed in Section \ref{conclusion}. Although TriCL is a universal framework that can be utilized in any type of trimodal system, this paper assessed the performance of TriCL on molecular graph representation learning tasks. We believe TriCL can be applied to other computer science tasks such as multilingual representation learning in the natural language processing field \citepApp{pan2021contrastive, wei2020learning} or video representation learning in the computer vision field \citepApp{sermanet2018time, xu2021videoclip}. Moreover, we believe TriCL also could be utilized in applied fields like biological representation learning through DNA-RNA-Protein (genome-transcriptome-proteome) central dogma. The wide application of TriCL in various fields would be an intriguing future work.
    
    \item \textit{Higher-modal Contrastive Learning} TriCL well performs in trimodal contrastive learning tasks by simultaneously contrasting triplet representations using Triangular Area Loss. As we devised a new loss function appropriate for trimodal tasks which are different from pairwise contrastive losses, a new form of losses would be required for the effective higher-modal contrastive learning task. For example, contrasting volumes of tetrahedrons would be a great starting point for the tetramodal system.
    
    \item \textit{Generative Tasks} In terms of molecular representation learning, molecular generation is an important research topic. Although this paper comprehensively discussed embedding spaces, molecular generation or optimization was not involved in the scope. Based on discriminated nature of each modality, it would be an interesting research to generate molecules having desired structural or sequential properties from the GNN embedding space.
\end{enumerate}

\subsection{Broader Impacts on Society}
\textbf{Potential Positive Impacts}
\begin{itemize}
    \item \textit{Understanding Cognitive Process} Multimodal contrastive learning in fact resembles the human cognitive process. As TriCL `simultaneously' contrasts three data representations, human collects diverse sensory information `as a whole' and forms virtual cognitive space by finding relations between them. We expect we could improve our understanding of human cognition by studying TriCL.
    \item \textit{Drug discovery} This study is not restricted to graph representation learning, but we extended pre-trained TriCL on the specific biological task in Section \ref{casestudy}. As we tested the power of pre-trained embedding space on GPCR-binding molecules, we expect that well-trained molecular embedding space could help develop new drugs for novel target proteins which require sophisticated structural design. Specifically, TriCL would be effective in developing drugs for rare diseases because TriCL could rapidly narrow down drug candidates only by using graph representations without a huge investment.
\end{itemize}

\textbf{Potential Negative Impacts}
\begin{itemize}
    \item \textit{Chemical Hazards} As a counter-effect of improved understanding of molecules, TriCL could be used in developing harmful chemicals. In fact, an uncontrolled dose of drugs also could role as a hazard. To prevent abuse, societal monitoring of chemical weapon development and consistent responsibilities on ethics of scientific knowledge and technology is required.
    \item \textit{Environmental Impact} Even though we have reduced the parameters and controlled the model size to our best, the process of training and implementing TriCL could still increase the carbon emissions slightly when compared with lighter unimodal framework. Developing a more efficient version of TriCL is among our considerations of future work.
\end{itemize}
{
\newpage 
\bibliographystyleApp{plainnat}
\bibliographyApp{appendixref}
}
\end{document}